\newcommand{\eqautoref}[1]{\hyperref[#1]{Eq.~(\ref*{#1})}}

\documentclass[times, review, 10pt]{elsarticle}
\usepackage{url}

\usepackage{amssymb}

\usepackage{xcolor}

\journal{kk}

\begin{document}

\begin{frontmatter}



\title{MuNet: A Mutualistic Network for Joint 3D Human Mesh Recovery and 3D Clothed Human Reconstruction from Single Images}

\author[label1,label2]{Yunqi Gao}

\author[label1,label2]{Leyuan Liu\corref{cor1}}

\author[label3]{Yuhan Li}

\author[label4]{Changxin Gao}

\author[label1,label2]{Jingying Chen}

\cortext[cor1]{Corresponding author}
\affiliation[label1]{
    orgdiv={National Engineering Research Center for E-Learning},
    organization={Central China Normal University},
    city={Wuhan},
    postcode={430079},
    state={Hubei},
    country={China}
}

\affiliation[label2]{
    orgdiv={National Engineering Research Center of Educational Big Data},
    organization={Central China Normal University},
    city={Wuhan},
    postcode={430079},
    state={Hubei},
    country={China}
}

\affiliation[label3]{
    orgdiv={School of Electronic Information and Communications},
    organization={Huazhong University of Science and Technology},
    city={Wuhan},
    postcode={430073},
    state={Hubei},
    country={China}            
}

\affiliation[label4]{
    orgdiv={School of Artificial Intelligence and Automation},
    organization={Huazhong University of Science and Technology},
    city={Wuhan},
    postcode={430073},
    state={Hubei},
    country={China}          
}



\begin{abstract}
3D human mesh recovery and 3D clothed human reconstruction are inherently related, yet they have long been studied in isolation, thereby overlooking the potential gains of joint optimization. To overcome this limitation, we propose to address these two tasks within a unified framework, which allows their mutual dependencies to be effectively exploited. Building on this idea, we propose MuNet, a mutualistic network for joint 3D human mesh recovery and 3D clothed human reconstruction from single images. First, we adopt 2-manifold graphs as a unified representation for all 3D models, enabling consistent modeling across 3D human mesh recovery and clothed human reconstruction. Second, we design an end-to-end graph convolutional network that progressively deforms an initial graph into a 3D human mesh and refines it into a detailed 3D clothed human model. Third, we introduce a mutualistic mechanism that allows reciprocal interaction between the two tasks {during training}, where 3D human mesh recovery provides guidance for 3D clothed human reconstruction, and reconstruction feedback refines the 3D human mesh recovery.
We extensively evaluate MuNet on six benchmark datasets for 3D human mesh recovery and 3D clothed human reconstruction, including Human3.6M, 3DPW, MPI-INF-3DHP, THuman2.0, CAPE, and RenderPeople. Experimental results demonstrate that MuNet achieves state-of-the-art performance on both tasks across all datasets.
The code of MuNet is released for research purposes at 
\url{https://github.com/starVisionTeam/MuNet}.
\end{abstract}


\end{frontmatter}

\begin{figure}[t]
\centering
\includegraphics[width=\textwidth]{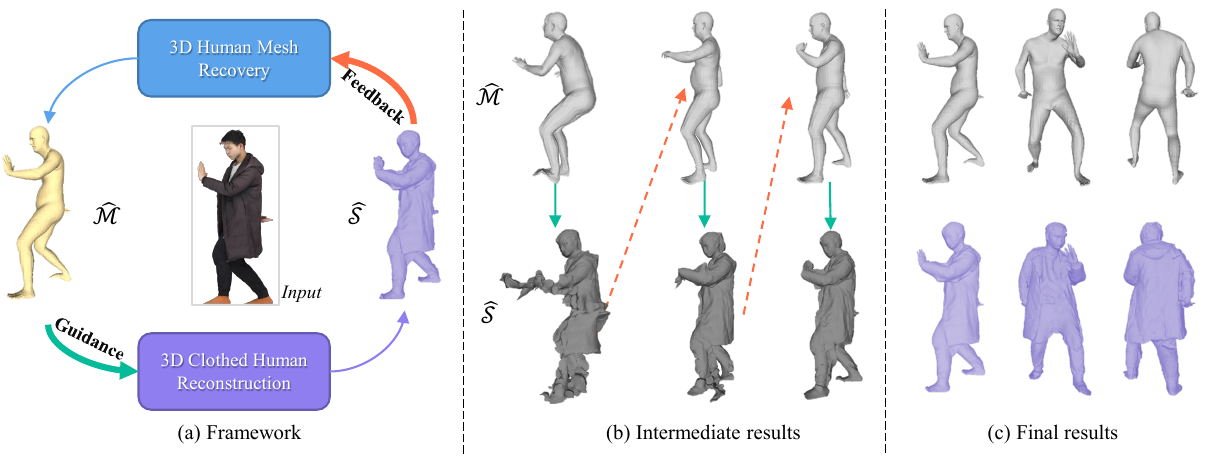}
\caption{(a) Framework of joint 3D human mesh recovery and 3D clothed human reconstruction. The tasks of 3D human mesh recovery and 3D clothed human reconstruction form a closed loop: the 3D body model ($\hat{\mathcal{M}}$) generated by 3D human mesh recovery serves as guidance for reconstructing the surface of the 3D clothed human ($\hat{\mathcal{S}}$). 
Subsequently, the surface reconstruction errors are fed back to the 3D human mesh recovery task for improving the 3D body model.
(b) Intermediate results generated by the two tasks during training. It can be observed that 3D human mesh recovery and  3D clothed human reconstruction are mutually beneficial. As with iterations, the results of both tasks are gradually improving. (c) The final results show that our method can recover accurate 3D human meshes of humans wearing loose-fitting clothing and reconstruct highly-fidelity 3D clothed human models.}
\label{fig:rets}
\end{figure}

\section{Introduction}
3D human mesh recovery and 3D clothed human reconstruction are two active topics in the computer vision and computer graphics communities. 
3D human mesh recovery, also referred to as 3D body model regression~\cite{ pymaf-X,pymaf} or 3D human pose and shape recovery/regression~\cite{HMR, SPIN}, concentrates on estimating the 3D poses and shapes of human bodies beneath clothing, whereas 3D clothed human reconstruction aims to reconstruct the 3D surfaces of humans with their garments.
These two tasks not only involve distinct problem definitions but also serve different application domains~\cite{survey}. 3D human mesh recovery is frequently utilized in applications such as virtual try-on and human motion analysis, whereas 3D clothed human reconstruction is commonly employed in film and game production, virtual reality, and the metaverse. Owing to the distinctness between 3D human mesh recovery and 3D clothed human reconstruction, existing studies typically treat them in isolation, thereby overlooking the opportunity to address their respective limitations and harness their mutual benefits through joint optimization.

Although notable advancements have been made in single-image 3D human mesh recovery in recent years~\cite{survey}, it still struggles with humans wearing various clothing. Hundreds of studies~\cite{SPIN, GCMR, DecoMR, pymaf, pymaf-X, tokenhmr, meshpose} have been dedicated to estimating 3D human poses and body shapes, typically represented by parametric SMPL-(X) models~\cite{SMPL, SMPL-X} or non-parametric 3D vertices. Essentially, these methods recover the 3D human mesh by aligning intermediate representations such as joints~\cite{HMR, pymaf, pymaf-X, SMPLify} (or joint heatmaps~\cite{ neural, Pavlakos}), silhouettes~\cite{Pavlakos}, and segmentation~\cite{neural, NBF} between the image representation space and the 3D model space. However, clothing, especially loose-fitting garments, often obscures key body parts, making it challenging to accurately extract intermediate representations of human bodies from images. Unfortunately, few methods specifically address how to mitigate the negative impact of clothing on 3D human mesh recovery~\cite{ISAIR2023, clothhmr}. 
Due to insufficient consideration of clothing's impact, most current 3D human mesh recovery methods struggle to adapt to clothing variations and thus only work well on humans wearing non-loose clothing.
While 3D human mesh recovery remains an open challenge, many methods~\cite{PaMIR,  ICON, DIF, GTA, econ, VS} have leveraged its outcomes to regularize 3D clothed human reconstruction.
As the 3D human mesh provides prior knowledge about human body structure and shape, it has the potential to enhance the stability and accuracy of 3D clothed human reconstruction.
However, existing 3D clothed human reconstruction methods typically employ \textbf{\textit{off-the-shelf}} 3D human mesh recovery models such as GCMR~\cite{GCMR}, PyMAF(-X)~\cite{pymaf,pymaf-X}, and  PIXIE~\cite{pixie}, which are still far from perfect. 
Experimental results~\cite{PaMIR,  econ} show that these 3D clothed human reconstruction methods are sensitive to  the quality of the regularized 3D 
human meshes. Consequently, the failures in 3D human mesh recovery methods usually result in 3D clothed human reconstruction failures~\cite{PaMIR, econ}.
To better exploit the regularization effect of 3D human mesh recovery on 3D clothed human reconstruction, it is essential to establish an end-to-end network that jointly optimizes both tasks.
Such a network enables customization of the 3D human mesh recovery model to meet the requirements of 3D clothed human reconstruction, thereby achieving optimal regularization.

Given the inherent connection between 3D human mesh recovery and 3D clothed human reconstruction, jointly optimizing them within a unified framework could contribute to mutual performance enhancements for both tasks. 
Based on this insight, we propose a framework to jointly recover 3D human meshes and reconstruct 3D clothed humans from single images.
As illustrated in Fig.~\ref{fig:rets}(a), the two tasks are organized into a closed loop: the 3D human mesh generated by 3D human mesh recovery serves as guidance for reconstructing the surface of the 3D clothed human.
Subsequently, the surface reconstruction results provide feedback on the 3D human mesh recovery task to improve the 3D human mesh estimate.
Specifically, we propose an end-to-end neural network, termed the mutualistic network (\textit{MuNet}), to implement the framework.
First, we employ a consistent representation for all 3D models involved in both tasks by formulating them as 2-manifold graphs.
Second, we design an end-to-end graph convolutional network to jointly optimize the two tasks by progressively deforming an initial graph into a 3D human mesh and a detailed 3D clothed human model.
Third, we establish a mutualistic mechanism between 3D mesh recovery and 3D clothed human reconstruction: the former supplies local feature localization and global topology priors to guide the latter, while the latter strengthens the former by feeding back cues derived from surface reconstruction errors.
As shown in Fig.~\ref{fig:rets}(b), the intermediate results of 3D human mesh recovery and 3D clothed human reconstruction indicate the mutual benefits between both tasks. With increasing iterations, the performance of both tasks progressively improves.
The final results presented in Fig.~\ref{fig:rets}(c) demonstrate that MuNet effectively recovers accurate 3D human meshes of humans wearing loose-fitting clothing and reconstructs high-fidelity 3D clothed human models.
Extensive experimental results on six benchmark datasets show that our MuNet outperforms current state-of-the-art (SOTA) methods in both 3D human mesh recovery and 3D clothed human reconstruction.
The code of MuNet is released for research purposes at 
\url{https://github.com/starVisionTeam/MuNet}.

The main contributions of this paper can be summarized as follows:

     (1) We propose a unified framework that addresses 3D human mesh recovery and 3D clothed human reconstruction jointly, thereby opening up the potential for mutual performance enhancement between the two tasks. To the best of our knowledge, this is the first attempt to propose a unified framework that jointly tackles the two tasks.
     
    (2) We design an end-to-end graph convolutional network that progressively deforms an initial graph into a 3D human mesh and a clothed 3D human model, enabling joint optimization between 3D human mesh recovery and 3D clothed human reconstruction by employing a unified graph representation for both tasks and establishing a task-aligned network.
    
    (3) We propose a mutualistic mechanism between body mesh recovery and clothed human reconstruction, enabling reciprocal performance improvements that were previously unattainable due to the tasks being treated independently.
    
    (4) Our MuNet achieves SOTA results on both tasks of 3D human mesh recovery and 3D clothed human reconstruction.


\section{Related Work}

\subsection{3D Human Mesh Recovery}

Since the release of statistical body models, methods for 3D human mesh recovery have gradually leveraged these models to estimate the shape and pose of the human body from a single image. Balan et al.~\cite{4270338} pioneered a method for estimating SCAPE~\cite{SCAPE} model parameters from images. Recently, SMPL~\cite{SMPL} has become the mainstream 3D human mesh recovery in academia. This success is attributed to SMPL's open-source nature and the rapidly growing community support, which includes ground-truth acquisition algorithm~\cite{SMPLify},~\cite{MoSh}, datasets with SMPL extended annotations~\cite{Human3.6M}~\cite{3dpw}~\cite{mpi-inf-3dhp}~\cite{CAPE}~\cite{renderpeople}~\cite{THUman2.0}, and landmark research works~\cite{HMR}~\cite{SPIN}~\cite{vibe}.

Current methods for 3D human mesh recovery~\cite{HMR,DecoMR,NBF,DSR,GCMR,HybrIK,pymaf,3DCrowdNet,pymaf-X,tokenhmr,meshpose,LIU2026112239,LUO2025111626} recover the 3D human mesh by aligning intermediate representations such as  2D/3D joints~\cite{HMR,pymaf,pymaf-X,SMPLify,Unite_the_People,PaMIR,ICON}, 2D heatmaps~\cite{Virtual_Markers,3DCrowdNet,neural,Pavlakos,STRAPS,Inter-part,Sim2real}, silhouettes~\cite{Pavlakos,Unite_the_People,PaMIR,ICON}, semantic segmentation~\cite{neural,NBF,densepose,meshpose}, depth information~\cite{hdnet} between the image representation space and the 3D model space. HMR~\cite{HMR} uses 2D/3D joints to supervise the output of human mesh parameters. SMPLify~\cite{SMPLify} iteratively fits the joints of the SMPL model to 2D joints. VirtualMarkers~\cite{Virtual_Markers} 2D-estimated joints. Unite the People~\cite{Unite_the_People} improves mesh alignment by enforcing consistency between reprojected mesh silhouettes and real image silhouettes. ICON~\cite{ICON} iteratively refines mesh reconstruction using body silhouette information. NBF~\cite{NBF} leverages semantic body segmentation as input to predict the human mesh. MeshPose~\cite{meshpose} improves estimation accuracy by applying weak DensePose~\cite{densepose} supervision to localize a subset of mesh vertices in 2D precisely. HDNet~\cite{hdnet} incorporates depth information to reconstruct clothed human bodies. 

In summary, mainstream 3D human mesh recovery methods rely heavily on intermediate representations for supervision. However, in real-world scenarios, humans often wear loose-fitting clothing, which occludes key body parts and makes it difficult to extract these intermediate representations from images accurately. Accurately annotating the ground-truth intermediate representations for humans wearing loose clothing is also challenging. {~\cite{ISAIR2023} proposes optimizing human mesh recovery through losses designed for clothed human reconstruction. These approaches offer effective solutions to improve reconstruction accuracy under occlusion caused by loose garments.}

\subsection{3D Clothed Human Reconstruction}
3D clothed human reconstruction methods~\cite{PaMIR,ICON,econ,DeepHuman,VS,SIFU,HUANG2024110758,NEUPANE2026112071,PAN2026112008} utilize 3D human mesh recovery models~\cite{GCMR,pymaf,pymaf-X} as prior information. For instance,
DeepHuman~\cite{DeepHuman} utilizes the SMPL model, estimated by SPIN~\cite{SPIN}, and semantically voxelizes it to extract 3D features. PaMIR~\cite{PaMIR} employs a voxelized version of the SMPL model, estimated using GCMR~\cite{GCMR}, to regularize free-form implicit functions. ICON~\cite{ICON} uses the SMPL model, estimated by both PyMAF~\cite{pymaf} and PIXIE~\cite{pixie}, to guide a detailed normal estimator and a visibility-aware implicit surface regressor. ECON~\cite{econ} treats the SMPL-X model, estimated by PyMAF-X~\cite{pymaf-X}, as a 'canvas' on which the estimated front and back surfaces are stitched together. VS~\cite{VS} stretches the SMPL-X model, estimated by PyMAF-X~\cite{pymaf-X}, to reconstruct the clothed human model. SIFU~\cite{SIFU} combines a side-view decoupling transformer with a 3D-consistent texture refinement pipeline, where the SMPL-X model, estimated by PyMAF-X~\cite{pymaf-X}, serves as the guiding model. These methods rely on off-the-shelf 3D human mesh recovery methods, yet these models are still far from perfect. 

Experimental results show that these 3D clothed human reconstruction methods are highly sensitive to the quality of the regularized 3D human meshes. To address this issue, during the testing phase, intermediate representations are used to iteratively refine the results of body mesh recovery during the testing phase.
PaMIR~\cite{PaMIR} minimizes reconstruction artifacts and 2D joints of clothed human bodies to optimize the SMPL parameters. ICON~\cite{ICON}, ECON~\cite{econ}, DIF~\cite{DIF}, VS~\cite{VS} and SIFU~\cite{SIFU} optimize SMPL/SMPL-X parameters by minimizing normal maps, silhouettes, and joint discrepancies. However, these intermediate representations are influenced by clothing, which leads to inaccuracies and, consequently, errors in body mesh recovery. As a result, in the clothed human reconstruction network, the matching between the body mesh and image features becomes inaccurate, leading to discrepancies in the reconstructed results. PaMIR~\cite{PaMIR}, ICON~\cite{ICON}, and DIF~\cite{DIF} are prone to issues like broken hands and feet, while ECON~\cite{econ} and SIFU~\cite{SIFU} may lead to leg bending, and VS~\cite{VS} may cause loss of detail. To maximize the regularization effect of 3D mesh recovery on clothed human reconstruction, an end-to-end network is needed to optimize both tasks together, ensuring optimal results.

\section{Method}

\begin{figure}[ht]
\centering
\includegraphics[width=\textwidth]{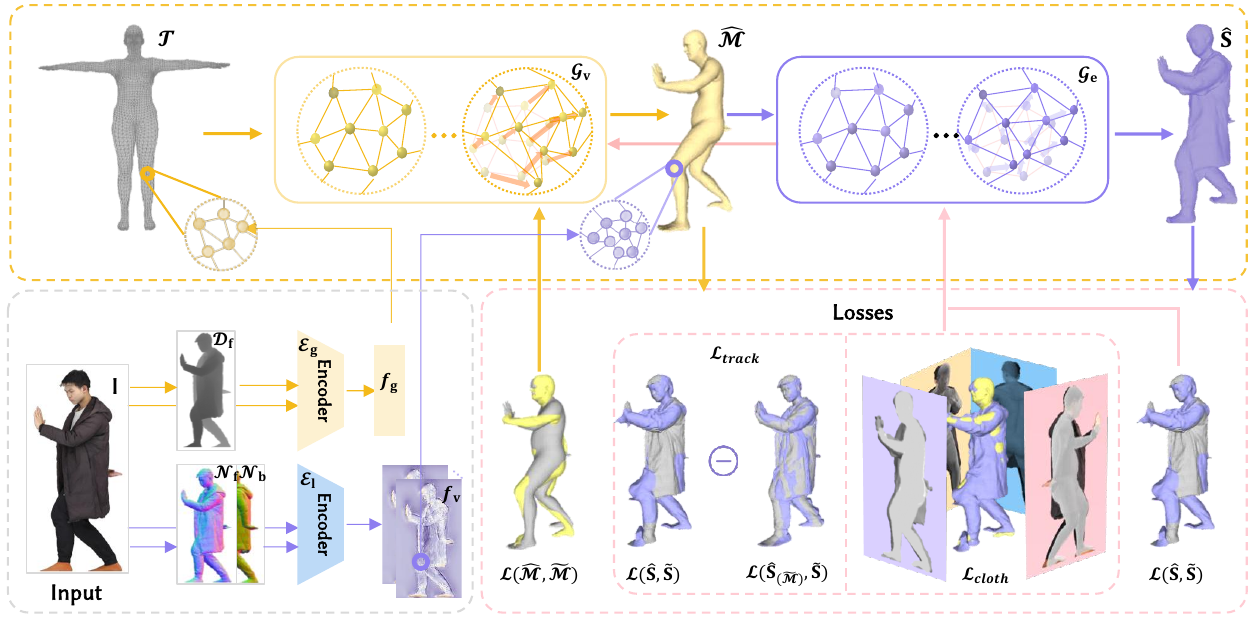}
\caption{Overview of MuNet. The T-pose SMPL template $\mathcal{T}$ is progressively deformed into a 3D human mesh $\hat{\mathcal{M}}$ and then refined into a clothed human model $\hat{\mathcal{S}}$ (yellow box).
This deformation process is guided by both global visual features $f_g$ and local pixel-aligned features $f_l$ (gray box).
To achieve better collaboration, MuNet establishes a mutual refinement mechanism: the body mesh provides structural guidance for clothing modeling, while clothing reconstruction refines the underlying mesh (pink box). }
\label{fig:overview}
\end{figure}

\subsection{Overview}

As the overview illustrated in Fig.~2, MuNet progressively deforms the initial 3D model ($\mathcal{T}$) into a 3D human mesh ($\hat{\mathcal{M}}$) and a 3D clothed human model ($\hat{\mathcal{S}}$). To this end, MuNet is designed as a graph convolutional network (GCN). To employ the GCN, all 3D models (i.e., $\mathcal{T}$, $\hat{\mathcal{M}}$, and $\hat{\mathcal{S}}$) involved are represented as graphs. Features derived from the input image are embedded into the graphs, enabling the GCN to deform the initial 3D model into a 3D human mesh and a 3D clothed human model consistent with the input image.
Built upon this representation, MuNet employs an end-to-end graph convolutional network to progressively deform an initial graph ($\mathcal{T}$) into a 3D human mesh ($\hat{\mathcal{M}}$) and further into a 3D clothed human model ($\hat{\mathcal{S}}$). 
MuNet also establishes a mutualistic mechanism between the tasks of 3D mesh recovery and 3D clothed human reconstruction: the former provides essential structural guidance to the latter, while the latter, in turn, refines and enhances the former. 

The model representation and feature embedding are presented in SubSec.~\ref{model_representation}, the network architecture of the GCN is elaborated in SubSec.~\ref{subsec:Architecture}, and the mutualistic mechanism between the two tasks is described in SubSec.~\ref{Mutualistic_Network}.

\subsection{Model Representation and Feature Embedding}
\label{model_representation}

Currently, 3D human mesh recovery methods and 3D clothed human reconstruction methods demonstrate inconsistencies in model representation. Models in 3D human mesh recovery methods are usually represented by the parametric SMPL-(X) model~\cite{HMR, SPIN, pymaf, pymaf-X, pixie,  HybrIK} or non-parametric 3D mesh vertices~\cite{GCMR, DecoMR, Virtual_Markers}, whereas models in 3D clothed human reconstruction methods are typically depicted by 3D voxel cube~\cite{DeepHuman,PaMIR} or implicit functions~\cite{Pifu,ICON}.
The inconsistent model representations make it difficult to deploy the two tasks into an end-to-end network and backpropagate gradients between the sub-networks of the two tasks, thereby hindering their unification. 

To address the inconsistent model representation issue, MuNet uniformly represents all 3D models in the two tasks, including the initial template ($\mathcal{T}$), the recovered 3D human mesh ($\hat{\mathcal{M}}$), and the reconstructed 3D clothed human model ($\hat{\mathcal{S}}$), by graphs. Specifically, we structure the graph of each 3D model into a 2-manifold triangular graph using~\cite{RWM}. In this way, each edge on a graph has four adjacent edges uniformly.
In addition to vertices and edges, we also embed visual features into the graphs of $\mathcal{T}$ and $\hat{\mathcal{M}}$.  Driven by these visual features, the initial template can be deformed into a 3D human mesh and a 3D clothed human model. Formally, the 3D models in MuNet are represented as:
\begin{eqnarray}\label{equ:stu}
\mathcal{T} &=& \left<\mathcal{V}_T, E, f_g\right> \\
\hat{\mathcal{M}} &=& \left<\mathcal{V}_M, E, \mathcal{F}\right> \\
\hat{\mathcal{S}} &=& \left<\mathcal{V}_S, E\right>
\end{eqnarray}
where $\mathcal{V}_T$, $\mathcal{V}_M$, $\mathcal{V}_S$ and  are vertices of $\mathcal{T}$, $\hat{\mathcal{M}}$, and $\hat{\mathcal{S}}$, $E$ is the edge set, $f_g$ and $\mathcal{F}$ are embedded visual features. 
As MuNet only alters the vertex positions of 3D models without adding or removing vertices and changing the edge connections, $|\mathcal{V}_T| = |\mathcal{V}_M| = |\mathcal{V}_S|$, and $\mathcal{T}$,$\hat{\mathcal{M}}$, and $\hat{\mathcal{S}}$ share the same edge set $E$.

We embed different visual features into the graphs of $\mathcal{T}$ and $\hat{\mathcal{M}}$ depending on the tasks they are associated with.
As mentioned above, the initial template $\mathcal{T}$ is deformed into a 3D human mesh $\hat{\mathcal{M}}$ that aligns with the body shape and the pose depicted in the input image.
The main goal of this task is to recover the global 3D topological structure of the human body. Therefore, we embed global visual features, extracted from the input image ($\mathcal{I}$) and the estimated depth map ($\mathcal{D}_f$) by the encoder $\mathcal{E}_g$, into each vertex of $\mathcal{T}$:
\begin{equation}    f_g = \mathcal{E}_{g}\left(\mathcal{I}, \mathcal{D}_f \right)
\end{equation}
Whereas $\hat{\mathcal{M}}$ is deformed by into a 3D clothed human model $\hat{{S}}$.
The goal of this task is to recover the local topological structure and details of the surfaces of the clothed human. To this end, we embed local visual features, extracted from the input image ($\mathcal{I}$) and the estimated normal maps ($\mathcal{N}_f$ and $\mathcal{N}_b$), into each vertex of $\hat{\mathcal{M}}$. We first generate feature maps $\mathcal{F}$ using the encoder $\mathcal{E}_l$:
 \begin{equation}
      \mathcal{F} =\mathcal{E}_{l}\left(\mathcal{I}, \mathcal{N}_f, \mathcal{N}_b \right)
 \end{equation}
Then, we embed local visual feature ($f_v$) on each vertex ($v$) of $\hat{\mathcal{M}}$:
\begin{equation}
{f_v} = {\cal U}\left( {{\cal F},v} \right)
\label{equ:fv}
\end{equation}
where $\mathcal{U}(\cdot,\cdot)$ is a local feature-puncturing function that takes features from the feature map $\mathcal{F}$. The concrete form of  $\mathcal{U}(\cdot,\cdot)$  will be introduced in Subsec.~\ref{Mutualistic_Network}.

\begin{figure}[tbp]
\centering
\includegraphics[width=\textwidth]{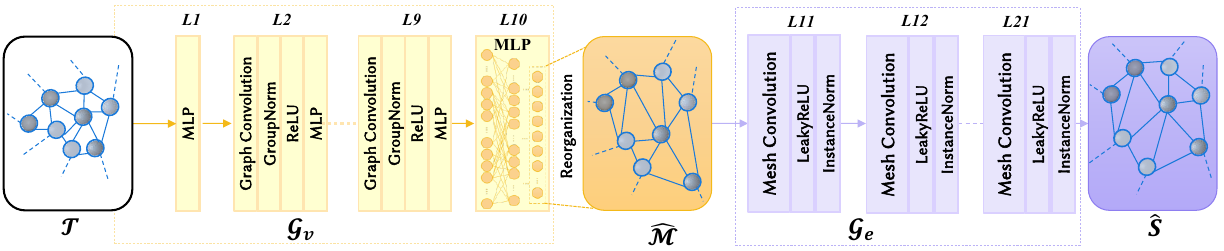}
\caption{Detailed network architecture of MuNet. The 3D mesh recovery sub-network $\mathcal{G}_v$ spans layers \textit{L1} to \textit{L10}, and the 3D clothed human reconstruction sub-network $\mathcal{G}_e$ spans layers \textit{L11} to \textit{L21}.}
\label{fig:architecture}
\end{figure}

\subsection{Network for Joint Tasks}

\label{subsec:Architecture}
The detailed network architecture of MuNet is illustrated in Fig.~\ref{fig:architecture}.
MuNet is designed as an end-to-end network that takes the initial graph ($\mathcal{T}$) as input and finally outputs the reconstructed 3D clothed human model ($\hat \mathcal{S}$).
The intermediate layer \textit{L10} of MuNet is designed to output 3D vertex positions of the deformed graph of $\mathcal{T}$. In our design, the network only updates vertex positions while implicitly preserving the edge connectivity inherent to $\mathcal{T}$. Consequently, the output vertices of $L10$ can be reorganized to yield the recovered 3D human mesh ($\hat{\mathcal{M}}$).
Beyond outputting the recovered 3D human mesh $\hat{\mathcal{M}}$, the reorganization of output vertices into a graph further facilitates mesh convolution operations in layers $\textit{L11} \sim \textit{L21}$.
Despite MuNet being an end-to-end network, we partition it into two sub-networks logically for clarity of presentation: the 3D human mesh recovery sub-network ($\mathcal{G}_v$, i.e., $\textit{L1} \sim \textit{L10}$) and the 3D clothed human reconstruction sub-network ($\mathcal{G}_e$, i.e., $\textit{L11} \sim \textit{L21}$). This partitioning is motivated by the observation that the layers before and after \textit{L10} are tailored to different objectives and utilize distinct graph convolution schemes.

\textbf{3D Human Mesh Recovery Sub-network.} The 3D body human mesh recovery sub-network aims to estimate the human body shape and pose from visual features, which requires capturing the human's global structure. 
To achieve this goal, in addition to embedding global features into the initial graph, we also design network architectures that can comprehensively explore global information.
Specifically, we start with a multilayer perceptron (MLP) layer ($L1$), enabling the network to aggregate global information from the initial graph. Then, a series of graph convolution layers ($L2 \sim L9$) is employed to deform the graph. 
Similar to GraphCNN~\cite{GraphCNN}, the graph convolution operation in each graph convolution layer is defined as:
\begin{equation}
{H^{'}} = \tilde A{H}{{\rm W} } 
\label{gcn1}
\end{equation}
where ${H}$ is the embedded features , ${\tilde A}$ is the adjacency matrix, and ${W}$ is the weight matrix.
This graph convolution operation can be interpreted as performing fully connected operations on each vertex, followed by an averaging operation over its neighborhood. Therefore, a GCN composed of such graph convolution operations can effectively capture the global topology of the human.
Finally, the sub-network terminates with an MLP layer ($L10$) designed to regress the 3D coordinates of each vertex on the deformed graph and the camera parameters.

\textbf{3D Clothed Human Reconstruction Sub-network.} 
The 3D clothed human reconstruction sub-network consists of a series of mesh convolution layers ($L11 \sim L21$).
This sub-network aims to reconstruct the surface of the clothed human by deforming the output of the 3D human mesh recovery sub-network (i.e., $\hat{\mathcal{M}}$). 
On one hand, $\hat{\mathcal{M}}$ has already captured the global topological structure of the human body; on the other hand, reconstructing the clothed human surface requires recovering local topological structures and surface details. Therefore, we design the 3D clothed reconstruction sub-network to focus on capturing local information.
To achieve this, the mesh convolution operation~\cite{meshcnn} is employed as the foundational structure for the 3D clothed human reconstruction sub-network. For each vertex on the graph, the mesh convolution operation only interacts with its directly adjacent vertices.
As previously stated, all graphs in MuNet are restructured into 2-manifold triangular graphs, where each edge ($e_0$) is exactly adjacent to four other edges ($e_1 \sim e_4$).
To leverage this stability for enhancing learning efficiency, we apply mesh convolution operations to the edges rather than the vertices. Therefore, the mesh convolution operation on the edge $e_0$ is defined as:
\begin{equation}
\sum\limits_{k = 0}^4 {\mu _k} \cdot {\psi}_{e_k} 
\label{gcn2}
\end{equation}
where ${\mu _k}$ is the weight of the convolution kernel, and  ${\psi}_{e_k}$ is the feature embedded on the edge ${e_k}$. Here, the feature embedded consists of the local visual features ($f_v$), the 3D coordinate ($v$), and the corresponding normal ($\mathcal{N}_{v}$):
\begin{equation}
    {\psi}_{{v}}=\left<f_v, {v}, \mathcal{N}_{{v}}\right>
    \label{gcn3}
\end{equation}
We concatenate the features embedded at the two vertices of edge ${e_k}$ and use it as the feature for the edge:
\begin{equation}
     {{{\psi}}_{{e_k}}} = {{{\psi}}_{{v_i}}} \oplus {{{\psi}}_{{v_j}}}
     \label{gcn4}
\end{equation}
where $v_i$ and $v_j$ are the two vertices of the edge $e_k$, and $\oplus$ denotes the concatenate operator.

\subsection{Mutualistic Mechanism}
\label{Mutualistic_Network}
A mutualistic mechanism is established between 3D human mesh recovery and 3D clothed human reconstruction. 3D human mesh recovery guides 3D clothed human reconstruction by providing geometry-aware local feature localization and global topological supervision, whereas 3D clothed human reconstruction refines the 3D human mesh through feedback from 3D surface reconstruction errors.


\textbf{Geometry-Aware Local Feature Localization.}
As mentioned earlier, local features are beneficial for reconstructing the surface details of clothed humans. However, in the absence of guidance from a 3D human mesh, directly localizing local visual features lacks a solid basis.
To this end, the localization of local features $f_v$ for 3D clothed human reconstruction relies on the geometric guidance of the 3D human mesh $\hat{\mathcal{M}}$.
A reasonable assumption is that the geometric position of each vertex on the reconstructed 3D clothed human surface lies in the vicinity of its corresponding vertex on the 3D human mesh. To this end, we extract local features from the neighborhood 
$\mathcal{P}$
of each vertex 
$v$ on the 3D human mesh $\hat{\mathcal{M}}$ for reconstructing the 3D clothed human:
\begin{equation}
f_v = \mathcal{U}\left( \mathcal{F}, \eta \left( \pi(v)\right) \right)
\end{equation}
where $\mathcal{F}$ denotes the feature map described in Equation~\ref{equ:fv}, $\pi(\cdot)$ is the weak perspective transformation that projects a vertex $v$ on the corresponding 3D point $p$ on the feature map $\mathcal{F}$,  $\eta(\cdot)$ is a neighborhood point sampling function that samples a point set $\mathcal{P}$ around $p$, and $\mathcal{U}(\cdot,\cdot)$ is a local feature-puncturing function that takes features from the feature map $\mathcal{F}$.
This geometry-aware feature extraction ensures spatial alignment between image cues and the 3D mesh, providing fine-grained local information for surface refinement. It also tightly couples the output of $\mathcal{G}_v$ with the input to $\mathcal{G}_e$, thereby enabling effective feature-level guidance across different stages of the network.

\textbf{Global Topological Supervision. } 
To ensure structural coherence between body and clothing reconstruction, we design a global topological supervision mechanism. The key idea is to enforce a shared mesh topology, so that the reconstructed human body mesh not only serves as a geometric prior but also directly guides the clothed surface recovery. Specifically, our framework first deforms a parametric 3D body model $\hat{\mathcal{M}}$ and then reconstructs the clothed surface $\hat{\mathcal{S}}$ in a sequential manner:
\begin{equation}
\mathcal{G}_e(\hat{\mathcal{M}}) \longrightarrow \hat{\mathcal{S}}
\end{equation}
By coupling the two stages under a unified topological constraint, the end-to-end reconstruction pipeline allows the body mesh to provide global structural guidance for clothing reconstruction while enabling each stage to leverage and enhance the outputs of the preceding one, promoting consistent human–clothing integration throughout the process.

\textbf{Surface Reconstruction Error Feedback.} 
To provide effective supervision for each sub-network and enable mutual feedback for collaborative optimization, MuNet designs three main loss functions.

For the 3D mesh recovery sub-network $\mathcal{G}_v$, the objective is to minimize the discrepancy between the predicted mesh $\hat{\mathcal{M}}$ and the ground-truth mesh $\tilde{\mathcal{M}}$. To achieve this, MuNet employs a composite loss function.
\begin{equation}
\mathcal{L}(\hat{\mathcal{M}},\rm{\tilde{\mathcal{M}} })=\mathcal{L}_{v}+\mathcal{L}_J+\mathcal{L}_{cd1}
\end{equation}
where $\mathcal{L}_{v}$ and $\mathcal{L}_{J}$ follow the vertex and joint supervision losses of ~\cite{HMR}, while the Chamfer distance loss $\mathcal{L}_{cd1}$ is adapted from ~\cite{VS} to further enforce geometric consistency.

For the 3D clothed human reconstruction sub-network $\mathcal{G}e$, MuNet employs a loss function defined on the predicted surface $\hat{\mathcal{S}}$ and the ground-truth surface $\tilde{\mathcal{S}}$.
\begin{equation}
\mathcal{L}(\hat{\mathcal{S}},\rm{\tilde{\mathcal{S}} })=\mathcal{L}_{cd2}+\mathcal{L}_n
\label{eq:L(ss)}
\end{equation}
Here, $\mathcal{L}_{cd2}$ denotes the Chamfer distance between $\hat{\mathcal{S}}$ and $\tilde{\mathcal{S}}$, while $\mathcal{L}_{n}$ represents a normal-based consistency loss, both like~\cite{VS}. 

To enhance the interaction between $\mathcal{G}_v$ and $\mathcal{G}_e$, MuNet introduces two collaborative losses, $\mathcal{L}_{\rm trace}$ and $\mathcal{L}_{\rm cloth}$.
The surface reconstruction error can be considered as the combination of (a) the inherent error of the surface reconstruction algorithm, and (b) the error introduced by SMPL guidance. 
In the loss defined in Equ.~\ref{eq:L(ss)}, the overall reconstruction error $\mathcal{L}(\hat{\mathcal{S}}, \tilde{\mathcal{S}})$ already includes the algorithmic error; thus, after discounting this inherent component, the remaining error primarily reflects the effect of SMPL guidance.
Since these components cannot be completely decoupled, $\mathcal{L}_{\rm trace}$ provides an approximate feedback of the SMPL-induced error. To compute it, the ground-truth SMPL model $\tilde{\mathcal{M}}$ is input to $\mathcal{G}_e$ to produce $\hat{\mathcal{S}}_{\tilde{\mathcal{M}}}$, and the difference between the overall reconstruction error and the reconstruction error using the ground-truth SMPL yields:
 
\begin{equation}
    \mathcal{L}_{\rm trace} = \mathcal{L}(\hat{\mathcal{S}}, \tilde{\mathcal{S}}) - \mathcal{L}(\hat{\mathcal{S}}_{\tilde{\mathcal{M}}}, \tilde{\mathcal{S}})
\end{equation}

The second collaborative loss, $\mathcal{L}_{\rm cloth}$, measures the discrepancy in clothing contours. The clothed human model is first subtracted by the corresponding human mesh, and the results are rendered from four canonical viewpoints $\{0^\circ, 90^\circ, 180^\circ, 270^\circ\}$. Denote the 2D projections of the clothed human model as $\{m_F^{\mathcal{S}*}, m_B^{\mathcal{S}*}, m_E^{\mathcal{S}*}, m_W^{\mathcal{S}*}\}$, and the 2D projections of the human mesh as $\{m_F^{\mathcal{M}*}, m_B^{\mathcal{M}*}, m_E^{\mathcal{M}*}, m_W^{\mathcal{M}*}\}$.
\begin{equation}
\mathcal{L}_{\rm cloth} = \Bigg| \sum_{k \in \{F,B,E,W\}} \big| m_k^{\hat{\mathcal{S}}} - m_k^{\hat{\mathcal{M}}} \big| - \sum_{k \in \{F,B,E,W\}} \big| m_k^{\tilde{\mathcal{S}}} - m_k^{\tilde{\mathcal{M}}} \big| \Bigg|
\end{equation}
Both $\mathcal{L}_{\rm trace}$ and $\mathcal{L}_{\rm cloth}$ contribute to the optimization of $\mathcal{G}_v$ and $\mathcal{G}_e$, ensuring coordinated training of the two sub-networks.
\begin{equation}
\mathcal{L}({\mathcal{S}}^*, {\mathcal{M}}^*) = \lambda_1 \, \mathcal{L}_{\rm trace} + \lambda_2 \, \mathcal{L}_{\rm cloth}
\end{equation}
{The collaborative losses provide joint supervisory signals for the two sub-networks during training, improving the consistency between body mesh recovery and clothed human reconstruction.}

\section{Experimental Results}
\subsection{Implementation Details}
In our implementation, ResNet-50 with the final fully connected layer removed serves as the encoder $\mathcal{E}_g$, and the encoder from VS~\cite{VS} is incorporated as the encoder $\mathcal{E}_l$. Adam is adopted as the network optimizer. MuNet is trained for 40 epochs, with a fixed batch size of 16. The learning rate is set to $3 \times 10^{-4}$ for the first 20 epochs and then decreases to $3 \times 10^{-5}$ for the remaining 20 epochs. The training takes about 9 days on a personal computer equipped with an NVIDIA GeForce RTX 4090 GPU.

\subsection{Datasets}

\begin{figure}[tbp]
\centering
\includegraphics[width=\textwidth]{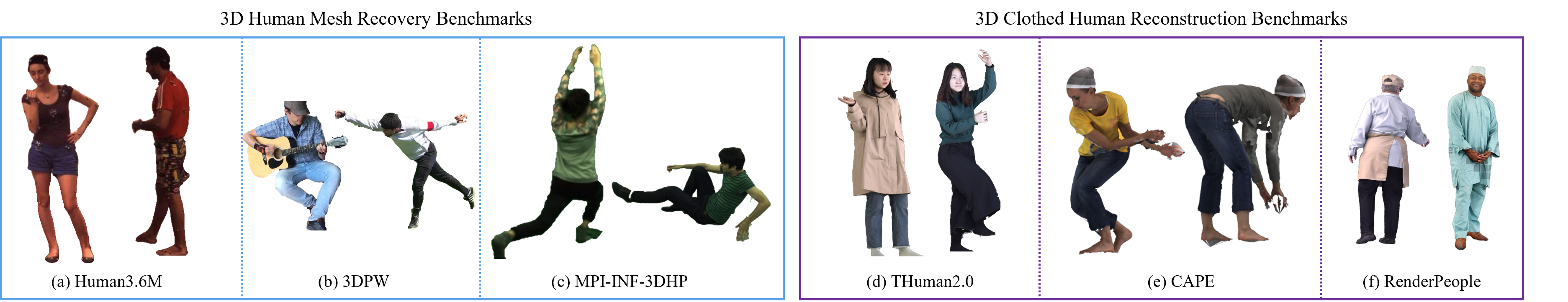}
\caption{Example images in six testing datasets. The testing datasets span diverse scenarios: Human3.6M, 3DPW, and MPI-INF-3DHP include structured indoor, in-the-wild, and non-standing poses, while CAPE, THuman2.0, and RenderPeople feature complex poses and varied clothing.}
\label{fig:datasets}
\end{figure}

\textbf{Training Dataset.} We train MuNet on the THuman2.0 dataset~\cite{THUman2.0}, as it provides both the registered SMPL fittings for 3D human mesh recovery and the ground-truth 3D human scans for 3D clothed human reconstruction. The data in THuman2.0 is captured from 525 college students featuring complex lab-acted poses and daily clothes. In our experiments, we randomly partition this dataset into training and testing sets using a 20:1 subject ratio. To augment the training data, we render 36 RGB images from each 3D human scan by rotating the scan with an interval of $10^o$ in the yaw angle. After data augmentation, we obtain a training dataset comprising 18K triplets of ⟨RGB image, SMPL fitting, 3D human scan⟩. To achieve more stable training results, we first pretrain $G_v$ on the training sets of Human3.6M~\cite{Human3.6M}, 3DPW~\cite{3dpw}, and MPI-INF-3DHP~\cite{mpi-inf-3dhp}.
The Human3.6M dataset (Fig.~\ref{fig:datasets} (a)) is a large-scale dataset with 3.6 million human poses in 17 activities performed by 17 professional actors in 17 different scenarios.
The 3DPW dataset (Fig.~\ref{fig:datasets} (b)) consists of 60 in-the-wild video sequences of 7 subjects engaging in various activities like playing guitar, sporting, and having coffee.
MPI-INF-3DHP (Fig.~\ref{fig:datasets} (c)) is comprised of images of 8 subjects with 2 types of clothing in both constrained indoor and complex outdoor scenes. Many people in the MPI-INF-3DHP dataset are in sitting, lying, and other non-standing poses, making this dataset rather challenging.

\textbf{Testing Datasets.} We evaluate our MuNet and other methods on three benchmark datasets for 3D human mesh recovery (i.e., Human3.6M~\cite{Human3.6M}, 3DPW~\cite{3dpw}, and MPI-INF-3DHP~\cite{mpi-inf-3dhp} testing set) and three benchmark datasets for 3D clothed human reconstruction (i.e., the THuman2.0~\cite{THUman2.0} testing set, CAPE~\cite{CAPE}, and RenderPeople~\cite{renderpeople}).
Fig.~\ref{fig:datasets} demonstrates some typical examples in the six testing datasets.
When evaluating 3D human mesh recovery methods, we use the standard testing sets of Human3.6M, 3DPW, and MPI-INF-3DHP, which respectively contain 363,928, 11,453, and 24,891 images.
CAPE ({Fig.~\ref{fig:datasets} (e)}) and RenderPeople ({Fig.~\ref{fig:datasets} (f)}) are two additional benchmarks, alongside THuman2.0 ({Fig.~\ref{fig:datasets} (d)}), commonly used for evaluating 3D clothed human reconstruction methods. 
The humans in CAPE  exhibit complex poses, and those in THuman2.0 and RenderPeople wear various loose clothing.
When assessing 3D clothed human reconstruction methods on the THuman2.0, CAPE, and RenderPeople datasets, we render six testing images from each 3D human scan at random viewpoints, yielding 150, 450, and 210 test images, respectively.


\subsection{Results on 3D Human Mesh Recovery}
We compare our MuNet with seven SOTA 3D human mesh recovery methods with publicly available code: SPIN~\cite{SPIN}, GCMR~\cite{GCMR}, DecoMR~\cite{DecoMR}, PyMAF~\cite{pymaf}, 
PyMAF-X~\cite{pymaf-X}, VirtualMaker~\cite{Virtual_Markers} and MeshPose~\cite{meshpose}. As re-training these methods on THuman2.0 resulted in performance degradation, we opted to use the official code and models to preserve optimal performance.  In addition to evaluating these methods on three 3D human mesh recovery benchmarks (i.e., Human3.6M~\cite{Human3.6M}, 3DPW~\cite{3dpw}, and MPI-INF-3DHP~\cite{mpi-inf-3dhp}), we also assess their performance on three 3D clothed human reconstruction benchmarks (i.e., THuman2.0~\cite{THUman2.0} testing set, CAPE~\cite{CAPE}, and RenderPeople~\cite{renderpeople}), as SMPL-(X) fittings are also available in these datasets.
\begin{table}
 \caption{Qualitative Comparisons of 3D human mesh recovery methods.}
\makebox[\textwidth]{\small (a) Quantitative comparison on the Human3.6M, 3DPW, and MPI-INF-3DHP datasets.} 
\resizebox{\linewidth}{!}{
\begin{tabular}{c|cc|ccc|cc}
\hline
  &   \multicolumn{2}{c|}{Human3.6M} &\multicolumn{3}{c|}{ 3DPW}&\multicolumn{2}{c}{MPI-INF-3DHP}\\
 Methods    & MPJPE & PA-MPJPE & MPJPE &  PA-MPJPE & MVPE & MPJPE & PA-MPJPE \\
\hline
SPIN~\cite{SPIN}       & 50.0 & 39.6 & 103.8 & 56.2 & 128.0	& 103.7 & 66.4\\

GCMR~\cite{GCMR}       & 52.5 & 44.9 & 122.2 & 66.2 & 137.4	& 132.7 & 77.0\\
DecoMR~\cite{DecoMR}    & 49.3 & 39.9 &	96.7 & 53.7 & 110.5	& 101.0 & 65.9\\
PyMAF~\cite{pymaf}     & 49.7 & 39.3 &  97.4 & 54.9 & 119.1 & 99.1  & 65.2 \\
PyMAF-X~\cite{pymaf-X}  & 47.7 & 37.0 & 78.0  &  47.1 &  91.3 & 93.5 & 56.5\\
VirtualMarker~\cite{Virtual_Markers} &47.3 & 32.0&67.5&41.3&77.9& 126.7&60.9\\
MeshPose~\cite{meshpose}&41.9&\textbf{29.4}&65.7&45.1&66.0&74.9&59.5\\
 \hline
 Ours  	& \textbf{38.2}& 32.4	& \textbf{49.2}& \textbf{38.9}& \textbf{54.4}   & \textbf{74.3}  & \textbf{56.3}\\
\hline
\end{tabular}
}

\vspace{1em}
\makebox[\textwidth]{\small (b) Quantitative comparison on the CAPE, THuman2.0, and RenderPeople datasets.}
\resizebox{\linewidth}{!}{
\begin{tabular}{c|ccc|ccc|cccc}
\hline
&  \multicolumn{3}{c|}{CAPE} & \multicolumn{3}{c|}{THuman 2.0} &\multicolumn{3}{c}{RenderPeople}\\
 Methods   & MPJPE & PA-MPJPE & MVPE &  MPJPE & PA-MPJPE & MVPE &MPJPE & PA-MPJPE  & MVPE\\
\hline
SPIN~\cite{SPIN}   &  74.2 &   52.4	&81.5 & 91.3 & 65.5 & 114.0	&  69.0 &44.4& 81.4&	\\

GCMR~\cite{GCMR}&67.7 & 54.5  	&79.0 &85.9&69.1&103.7	& 	57.1&46.5&67.7	\\

DecoMR~\cite{DecoMR}  &79.4&53.7 &   85.0	& 90.9&65.6 &111.6& 69.4 &46.3& 73.8	\\
PyMAF~\cite{pymaf}&75.0 &  51.0 &77.2	&87.5&63.0&108.7 	&  70.4& 44.8	&79.8\\
   
PyMAF-X~\cite{pymaf-X}&70.8 &   47.9	&   72.7 &84.2&58.2&96.0&66.4& 44.6&70.3\\
VirtualMarker~\cite{Virtual_Markers} & 59.9&44.6& 64.8&75.1 & 56.7& 92.0& 58.4&43.1&65.0 \\
MeshPose~\cite{meshpose}&68.1&48.2&79.1&69.9&50.8&83.6&60.3&41.6&67.3\\
   \hline
   Ours  &\textbf{53.3} &\textbf{41.1} &\textbf{60.0}  & \textbf{45.7}  &  \textbf{36.1} & \textbf{49.9}	&\textbf{42.3} & \textbf{33.0}& \textbf{48.6}	\\


\hline
\end{tabular}
}
\label{tab:cmp_body}
\end{table}
\subsubsection{Quantitative Comparisons}

\textbf{Metrics.} The \textit{Mean Per Joint Position Error} (MPJPE), \textit{Procrustes-aligned MPJPE} (PA-MPJPE), and \textit{Mean Per Vertex Error} (MVPE) are adopted as the quantitative metrics for evaluating 3D human mesh recovery methods. We report all three metrics on the 3DPW,  THuman2.0, CAPE, and RenderPeople datasets, while we only report MPJPE and PA-MPJPE on MPI-INF-3DHP and Human3.6M since the SMPL fittings are not provided in these two datasets.

The quantitative comparisons between our MuNet and other 3D mesh recovery methods are presented in Table~\ref{tab:cmp_body}.
As shown in Table~\ref{tab:cmp_body} (a), MuNet outperforms other methods in most metrics on the Human3.6M, 3DPW, and MPI-INF-3DHP benchmark datasets except for PA-MPJPE on the Human3.6M dataset, where it slightly lags behind VirtualMarkers and 
MeshPose. Particularly, MuNet surpasses other methods significantly in MPJPE and MVPE, indicating that MuNet not only estimates human pose and shape more accurately but also regresses human body scale and camera parameters more precisely.
As shown in Table~\ref{tab:cmp_body} (b), MuNet outperforms other methods across all metrics on the CAPE, THuman2.0, and RenderPeople datasets. The advantage of MuNet is more pronounced on the THuman2.0 and RenderPeople datasets. This is because these two datasets contain many images of humans wearing loose clothing, and MuNet excels at estimating the pose and shape of the human body beneath loose clothes, while other methods are less effective in this scenario.

\subsubsection{Qualitative Comparisons}
\begin{figure}[ht]
\centering
\includegraphics[width=\textwidth]{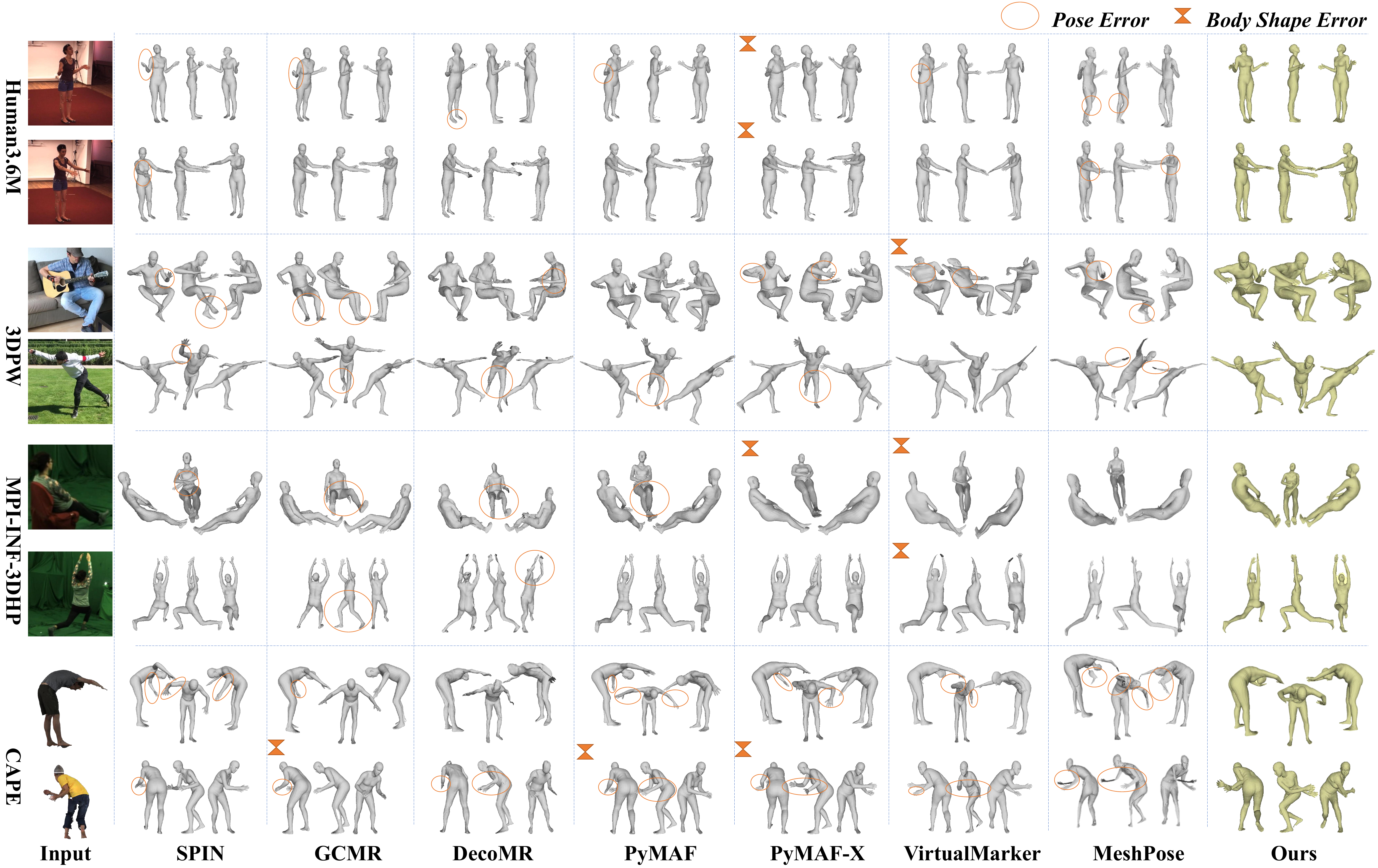}
\caption{Qualitative comparisons of different 3D human mesh recovery methods on the  Human3.6M, 3DPW, MPI-INF-3DHP, and CPAE datasets, where humans wear non-loose clothing. We show the results from three viewpoints and highlight the obvious pose and body shape errors visible from each viewpoint. Please zoom in to see the details.}
\label{fig:3DMesh_benchmark}
\end{figure}

\begin{figure}[ht]
\centering
\includegraphics[width=\textwidth]{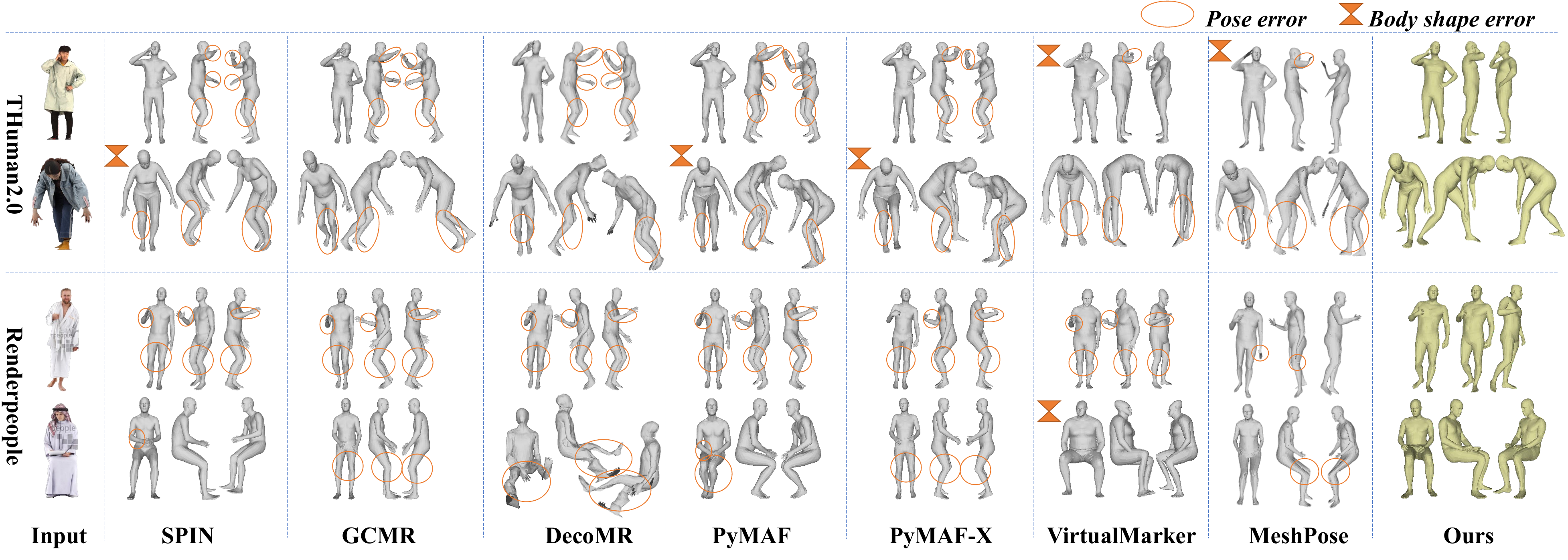}
\caption{Qualitative comparison of different 3D human mesh recovery methods on the THuman2.0 and RenderPeople datasets, where humans wear loose clothing. We show the results from three viewpoints and highlight the obvious pose and body shape errors visible from each viewpoint. Please zoom in to see the details.}
\label{fig:3dMesh_clothed}
\end{figure}

We divide the qualitative comparisons on the seven datasets into two groups based on whether the people in the images wear loose clothing. 
The results on Human3.6M, 3DPW, MPI-INF-3DHP, and CAPE, where humans wear non-loose clothing, are shown in Fig.~\ref{fig:3DMesh_benchmark}.
It can be observed from the results shown that most methods can accurately recover 3D human meshes for humans with relatively simple poses in the Human3.6 dataset. However, other methods struggle with pose or body shape errors for humans with more complex poses. In contrast, our MuNet can recover both the pose and body shape accurately.
The results on THuman2.0 and RenderPeople, where humans wear loose clothing, are shown in Fig.~\ref{fig:3dMesh_clothed}.
Methods like SPIN~\cite{SPIN}, GCMR~\cite{GCMR}, DecoMR~\cite{DecoMR}, PyMAF~\cite{pymaf}, and PyMAF-X~\cite{pymaf-X} incorrectly estimate the straight legs as bent, since they do not consider the concealment of loose clothing. Additionally, because of the influence of loose clothing, these methods also estimate the pose and body shape of the limbs incorrectly.
Conversely, MuNet excels at estimating the pose and shape of the human body beneath loose clothes because MuNet fully considers the impact of loose clothing on 3D human recovery.

\subsection{Results on 3D Clothed Human Reconstruction}

We compare our MuNet with four SOTA 3D clothed human reconstruction methods with publicly released code: PaMIR~\cite{PaMIR}, ICON~\cite{ICON}, ECON~\cite{econ}, D-IF~\cite{DIF}, VS~\cite{VS} and SIFU~\cite{SIFU}. We use the improved PaMIR code provided by Xiu et al.~\cite{ICON} to ensure that all methods involved in our comparison equally take the original RGB image and normal maps as input. 
All methods involved in our comparison employ recovered 3D human meshes represented by SMPL-(X) as guidance for reconstructing 3D clothed human models.

\begin{table}
 \caption{Qualitative Comparison of 3D clothed human reconstruction methods.}
\resizebox{\linewidth}{!}{
\begin{tabular}{c|c|ccccc|ccccc|ccccc}
\hline
  &  &\multicolumn{5}{c|}{CAPE} & \multicolumn{5}{c|}{THuman 2.0} &\multicolumn{5}{c}{RenderPeople}\\
 Methods   & SMPL(-X) Estimator & $\varepsilon_{cd}$ & $\varepsilon_{p2s}$ & $\varepsilon_{s2p}$ &  $\varepsilon_{cos}$  & $\varepsilon_{l2}$& $\varepsilon_{cd}$  & $\varepsilon_{p2s}$& $\varepsilon_{s2p}$ &  $\varepsilon_{cos}$  & $\varepsilon_{l2}$ & $\varepsilon_{cd}$    & $\varepsilon_{p2s}$ & $\varepsilon_{s2p}$& $ \varepsilon_{cos}$  & $ \varepsilon_{l2}$\\
\hline

PaMIR~\cite{PaMIR}   &GCMR~\cite{GCMR}& 6.352&4.165 &8.538 &0.3276 & 0.6569&4.170&3.238&5.103&0.3258&0.6737&4.080&3.291&4.866&0.3167 	&0.6561\\

ICON~\cite{ICON}&PyMAF~\cite{pymaf}&3.727&2.218&5.236&0.1886&0.6331&3.094&2.594&	3.595& 0.2481&0.8167	&2.712&	2.126&3.298&0.2127&0.7159\\
ECON~\cite{econ} &PyMAF-X~\cite{pymaf-X}&2.751&2.161	&3.342&0.1774&0.3931&2.295	&2.173&2.417& 0.2054	&0.4582		&2.345&	1.964&	2.727&0.1910&0.4278\\
D-IF~\cite{DIF}  &PyMAF~\cite{pymaf}&4.139&\textbf{2.152}&6.128&0.1877&0.6281&3.295&	2.661&3.929&0.2519	&0.8275&2.980&2.199&3.7611&0.2175	&0.7249\\
VS~\cite{VS}&PyMAF-X~\cite{pymaf-X}&4.348&2.944&5.752&0.2461&0.5170&4.437&3.204&5.670&0.3182&0.6590&3.658&2.765&4.552&0.2753&0.5849\\
SIFU~\cite{SIFU}&PyMAF~\cite{pymaf}&3.780&2.832&4.729&0.2237&0.7367&3.591&2.994&4.188&0.2780&0.9089&3.011&2.462&3.559&0.2336&0.7866\\
\hline
     Ours  &  ---& \textbf{2.430}	& 2.393	& \textbf{2.467}		&\textbf{ 0.1732}& \textbf{0.3247}  &\textbf{2.141}	& \textbf{2.093}	& \textbf{2.189}		&\textbf{0.2026}& \textbf{0.4299} 	& \textbf{1.345} & \textbf{1.236}&\textbf{1.455}&\textbf{0.1231}&\textbf{0.3049}\\
\hline
\end{tabular}
 \label{tab:chr_cmp}
}      
\end{table}

\begin{figure}[ht]
\centering
\includegraphics[width=\textwidth]{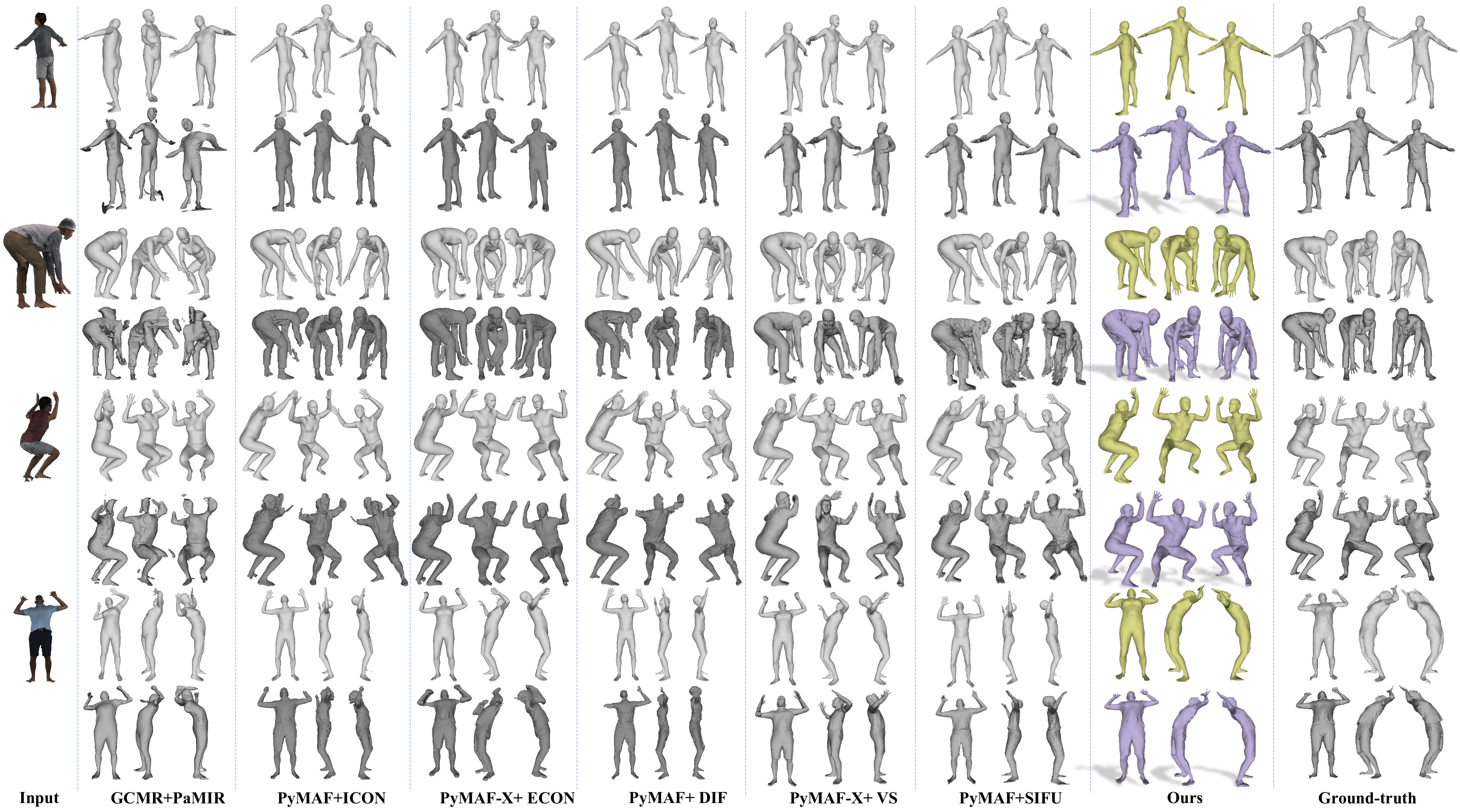}
\caption{Qualitative comparison of different 3D clothed human reconstruction methods on the CAPE dataset. For each input image, we show the SMPL-(X) recovered 3D human mesh and the reconstructed 3D clothed human models respectively in the even and odd rows. Please zoom in to see the details.}
\label{fig:clothed_CAPE}
\end{figure}

\begin{figure}[ht]
\centering
\includegraphics[width=\textwidth]{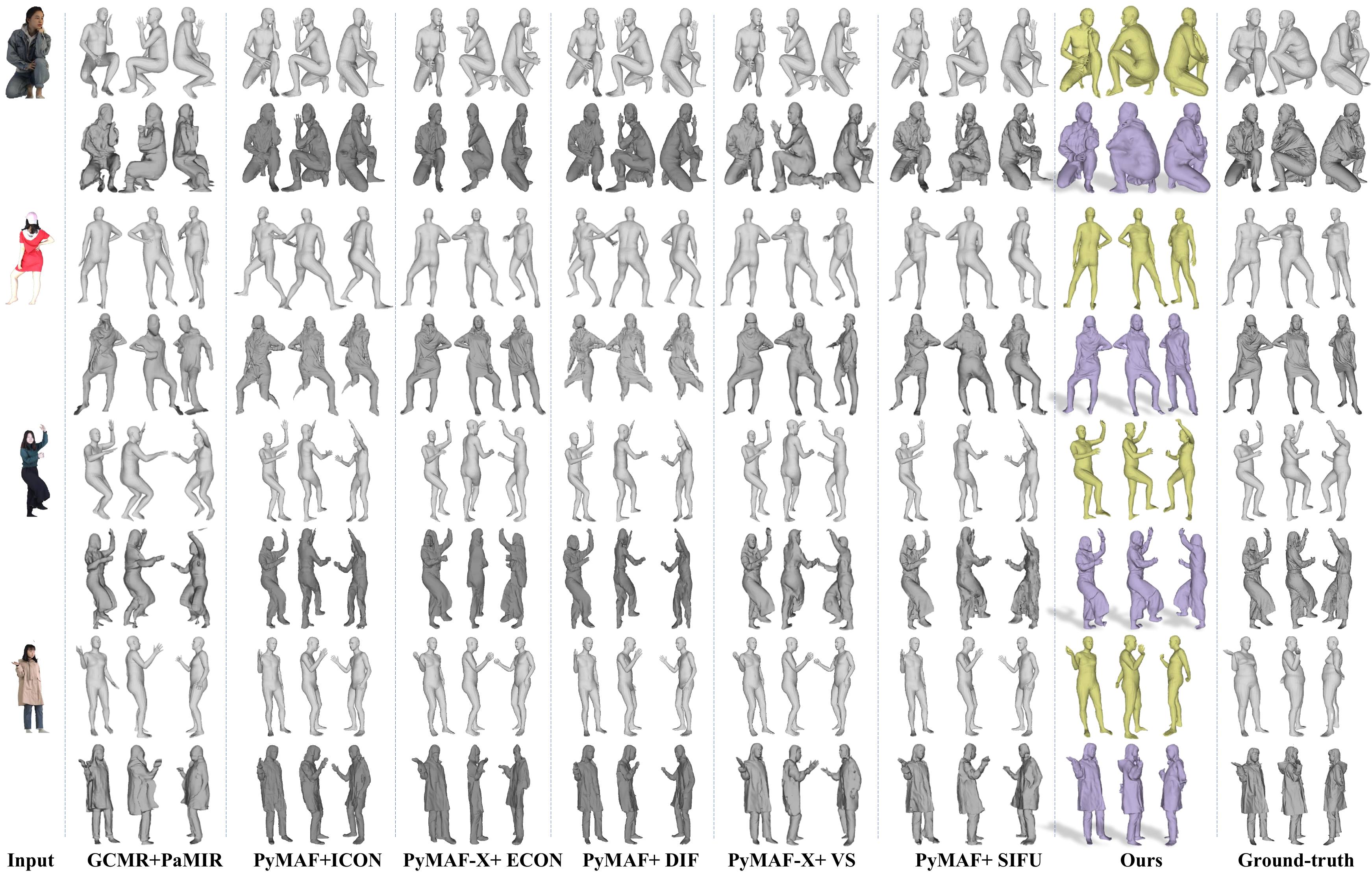}
\caption{Qualitative comparison of different 3D clothed human reconstruction methods on the THuman2.0 dataset. For each input image, we show the recovered 3D human meshes and the reconstructed 3D clothed human models respectively in the odd and even rows. Please zoom in to see the details.}
\label{fig:clothed_THuman2}
\end{figure}

\begin{figure}[ht]
\centering
\includegraphics[width=\textwidth]{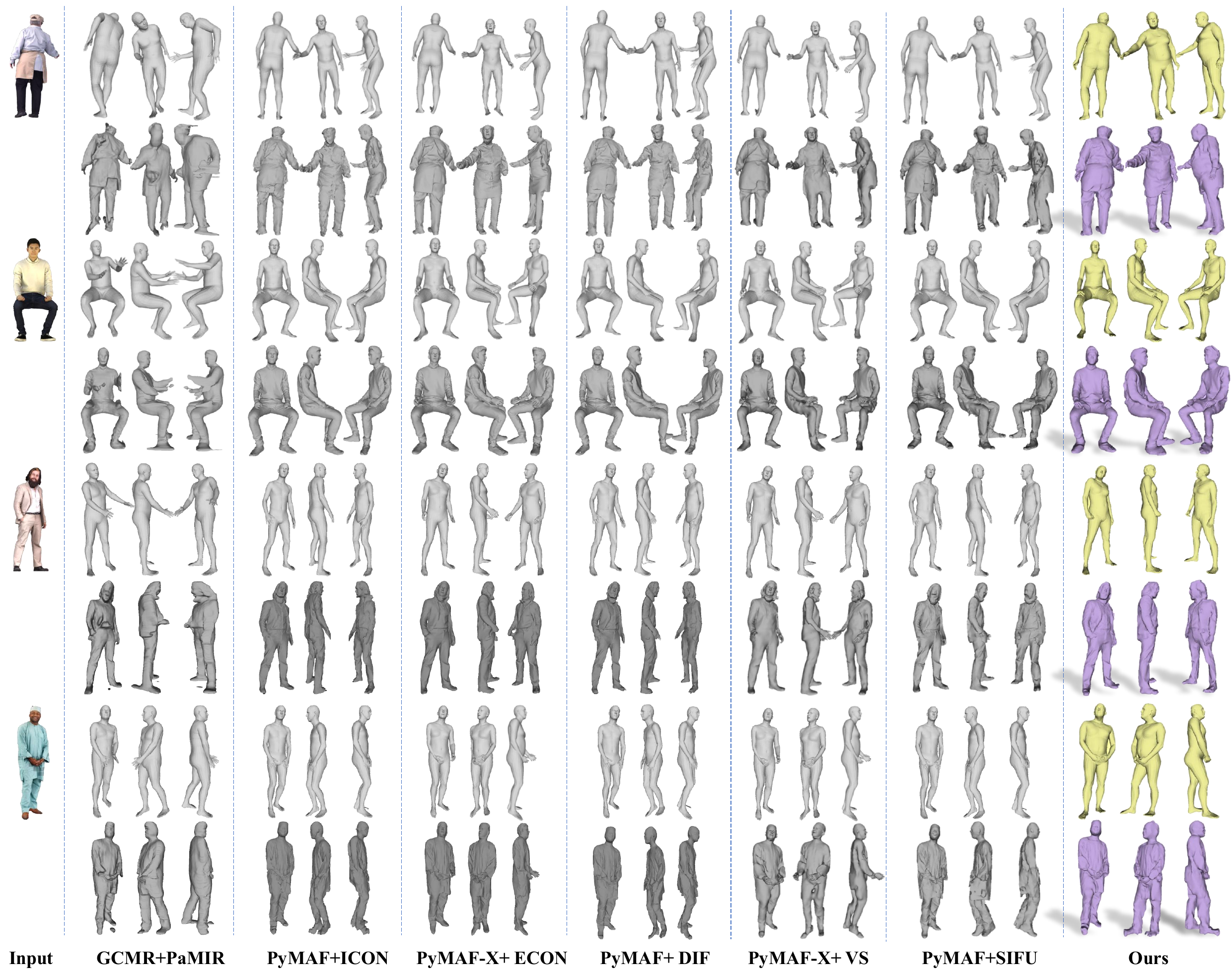}
\caption{Qualitative comparison of different 3D clothed human reconstruction methods on the RenderPeople dataset. For each input image, we show the recovered 3D human meshes and the reconstructed 3D clothed human models, respectively in the odd and even rows. Please zoom in to see the details.}
\label{fig:clothed_renderpeople}
\end{figure}

\subsubsection{Quantitative Comparisons}  

\textbf{Metrics.} \textit{Chamfer distance} ($\varepsilon_{cd}$), \textit{point-to-surface distance} ($\varepsilon_{p2s}$), \textit{surface-to-point distance} ($\varepsilon_{s2p}$), \textit{cosine distance} ($\varepsilon_{cos}$),
and \textit{L2 normal error} ($\varepsilon_{l2}$) are used as quantitative metrics for evaluating different 3D clothed human reconstruction methods.
The first two metrics mainly assess the global topology of reconstructed 3D clothed humans, while the latter two tend to evaluate local surface details.
As with ECON~\cite{econ}, we render four normal maps from both the ground-truth and reconstructed models at view angles of $\{0^o, 90^o, 180^o, 270^o\}$ for calculating the cosine distance and L2 normal error.

The quantitative comparisons between our MuNet and other SOTA methods are presented in Table~\ref{tab:chr_cmp}. Note that the results reported in the original papers of PaMIR~\cite{PaMIR}, ICON~\cite{ICON}, ECON~\cite{econ}, D-IF~\cite{DIF}, VS~\cite{VS} and SIFU~\cite{SIFU} are based on ground-truth SMPL-(X) and normal maps. However, SMPL-(X) and normal maps need to be estimated in practical applications and their accuracy significantly impacts the results of 3D clothed human reconstruction. Therefore, when reporting results in Table~\ref{tab:chr_cmp}, we use estimated SMPL-(X) and normal maps for all methods. 
As illustrated in Table~\ref{tab:chr_cmp}, MuNet surpasses all other SOTA 3D clothed human reconstruction methods across all metrics on all three benchmark datasets.
As illustrated in Table~\ref{tab:chr_cmp}, the results of all 3D clothed human reconstruction methods on the CAPE dataset are worse than those on the THuman2.0 and RenderPeople datasets. Recalling Table~\ref{tab:cmp_body}, it can be observed that the corresponding SMPL-(X) estimators in Table~\ref{tab:chr_cmp} also produce worse 3D human mesh recovery results on the CAPE dataset than those on the other two datasets. The results of MuNet are more stable than those of other methods, avoiding the low $\varepsilon_{p2s}$ and high $\varepsilon_{s2p}$ issues seen in methods such as ICON and DIF. For the THuman2.0 and Renderpeople datasets, MuNet surpasses all other SOTA 3D clothed human reconstruction methods in all metrics in these two benchmark datasets. Other methods do not recover the human mesh model accurately under loose clothing, leading to poor reconstruction of clothing details (as indicated by higher $\varepsilon_{cos}$ and $\varepsilon_{l2}$).
It indicates that the accuracy of 3D mesh recovery is crucial for 3D clothed human reconstruction.

\subsubsection{Qualitative Comparisons}  
The qualitative comparisons among different 3D clothed human reconstruction methods on the CAPE, THuman2.0, and RenderPeople datasets are respectively illustrated in Fig.~\ref{fig:clothed_CAPE}, Fig.~\ref{fig:clothed_THuman2}, and Fig.~\ref{fig:clothed_renderpeople}.
The humans in the CAPE dataset exhibit complex poses. If there are errors in the poses estimated by the SMPL-(X) estimators, 3D models reconstructed by 3D clothed human reconstruction methods like PaMIR~\cite{PaMIR}, ICON~\cite{ICON} DIF~\cite{DIF}, and VS~\cite{VS} are prone to produce artifacts such as disembodied limbs and non-human shapes.
In contrast, ECON~\cite{econ} and our MuNet are robust to pose variations.
Many individuals wear loose casual clothing in the THuman2.0 and RenderPeople datasets. In this scenario, PaMIR~\cite{PaMIR}, ICON~\cite{ICON}, and DIF~\cite{DIF} still suffer from artifacts, and PaMIR~\cite{PaMIR}, ICON~\cite{ICON}, DIF~\cite{DIF} ECON~\cite{econ}, and SIFU~\cite{SIFU} may produce incomplete clothes. In contrast, our MuNet demonstrates excellent generalization to loose clothing.

\subsection{Ablation Study}

\subsubsection{Effectiveness of 3D Clothed Human Reconstruction Sub-network}

\begin{figure}[h]
\centering
\includegraphics[width=0.7\textwidth]{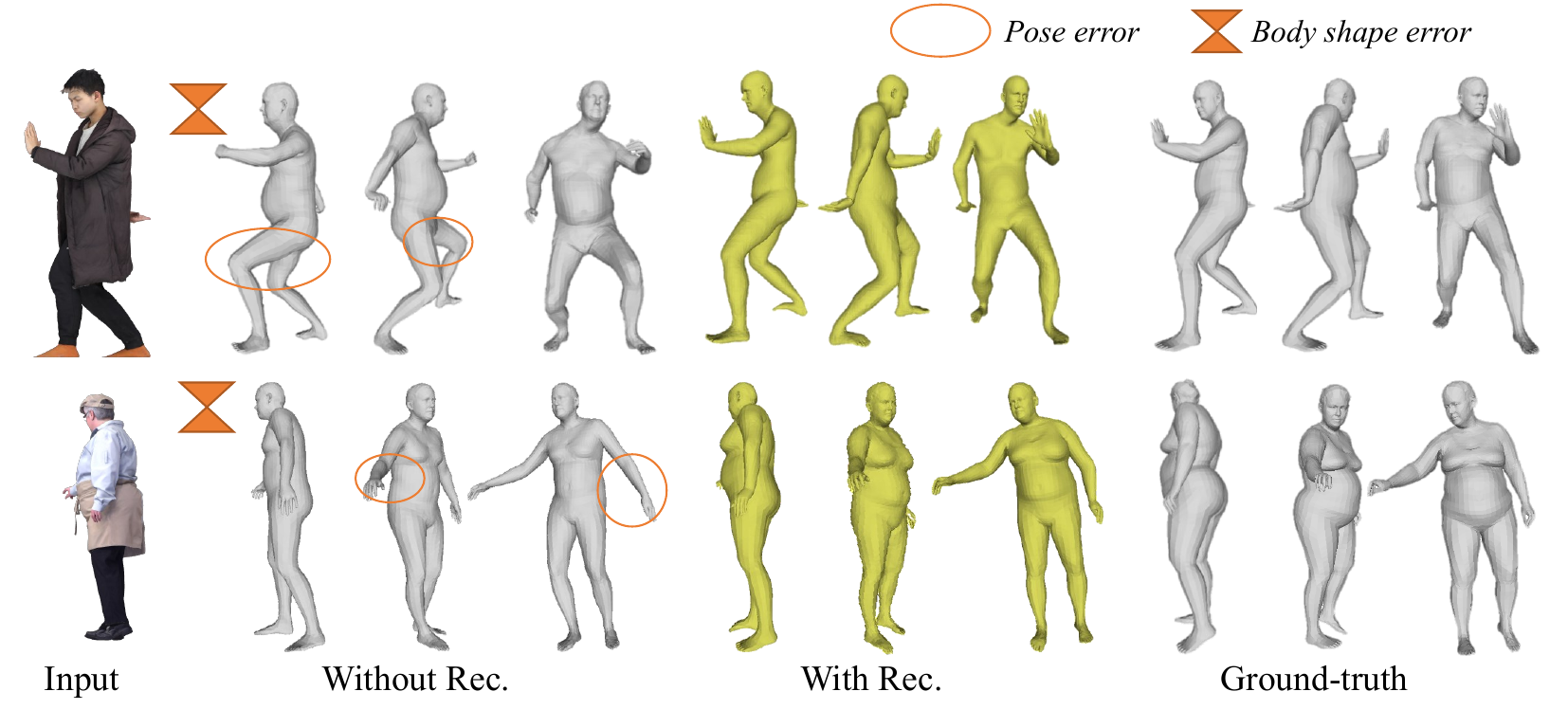}
\caption{Comparisons of 3D human mesh recovery results with and without clothed human reconstruction are shown. The first row presents pose errors under loose clothing, while the second row displays body shape errors under occlusions with adornments. Please zoom in to see the details.}
\label{ab1.pdf}
\end{figure}

To evaluate the effectiveness of the 3D clothed human reconstruction module, we implemented a variant that removes the 3D clothed human reconstruction module. As shown in Table~\ref{tab:abl1}, on the THuman2.0 dataset, incorporating the 3D clothed human reconstruction module results in reductions of MPJPE, PA-MPJPE, and MVPE by 31.7\%, 30.4\%, and 35.6\%, respectively. On the CAPE dataset, the errors in MPJPE, PA-MPJPE, and MVPE decrease by 8.1\%, 12.9\%, and 21.0\%, respectively. For the Renderpeople dataset, the reduction in these error metrics is 29.3\%, 30.0\%, and 24.1\%. Since the clothing in the CAPE dataset is more form-fitting, the reduction in metrics is smaller compared to the other two datasets. These results indicate that incorporating 3D clothed human reconstruction significantly improves the outcomes of 3D human mesh recovery. The visual results in Figure~\ref{ab1.pdf} show that removing the 3D clothed human reconstruction module leads to significant pose and body shape errors. For loose clothing occlusions, estimating the human pose becomes more challenging; while for occlusions with adornments, estimating the body shape is even more difficult. This further validates the conclusions drawn from Table~\ref{tab:abl1}.

\begin{table}
\caption{Impact of 3D clothed human reconstruction on human mesh recovery}
\resizebox{\linewidth}{!}{
\begin{tabular}{l|ccc|ccc|ccc}
\hline
  &\multicolumn{3}{c|}{THuman 2.0} & \multicolumn{3}{c|}{CAPE} &\multicolumn{3}{c}{Renderpeople}\\
 Methods    & MPJPE & PA-MPJPE & MVPE &  MPJPE & PA-MPJPE & MVPE &MPJPE & PA-MPJPE & MVPE\\
\hline
     w/o  Reconstruction   & 80.2  &  59.1 & 90.7	&67.6 & 51.4 & 72.6	&59.7& 44.9&  65.3	\\
    w/ Reconstruction   & \textbf{45.7}  &  \textbf{36.1} & \textbf{49.9}	&\textbf{53.3} &\textbf{41.1} &\textbf{60.0} &\textbf{42.3} & \textbf{33.0}& \textbf{48.6}	\\
\hline
\end{tabular}
\label{tab:abl1}
}
\end{table}

\subsubsection{Effectiveness of 3D Mesh Recovery Sub-network }
To evaluate the effectiveness of adaptive 3D human mesh recovery, we compared the results of reconstruction guided by other advanced methods, as shown in Table~\ref{tab:abl2}. 
As shown in the table, the clothed human reconstruction results using other advanced methods~\cite{SPIN,GCMR,DecoMR,pymaf,pymaf-X,Virtual_Markers} are inferior to those obtained with adaptive 3D human mesh recovery.
Our 3D human mesh recovery network is simpler than those in ~\cite{SPIN, DecoMR,pymaf,pymaf-X}, and it uses lower-dimensional image features compared to ~\cite{pymaf,pymaf-X, Virtual_Markers}. Nevertheless, using adaptive 3D human meshes for clothed human reconstruction still outperforms other off-the-self methods.

\begin{figure}[h]
\centering
\includegraphics[width=\textwidth]{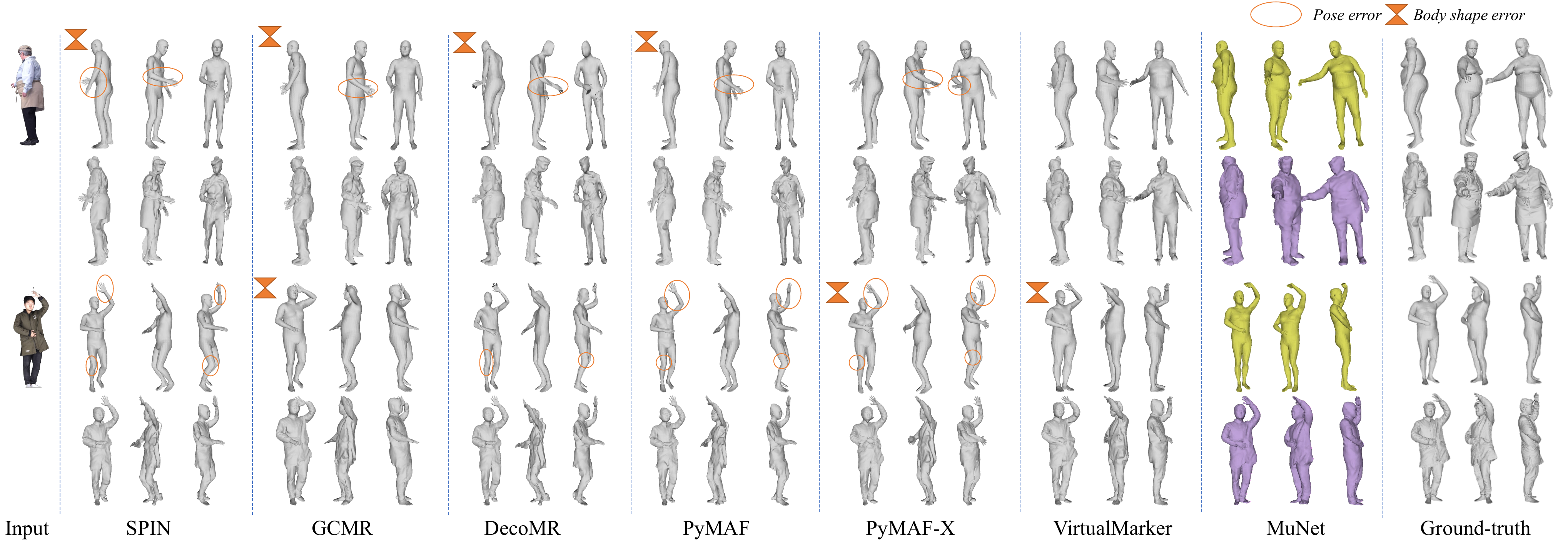}
\caption{The figure shows the results of clothed human reconstruction guided by advanced methods. The first data shows that ~\cite{SPIN,GCMR,DecoMR,pymaf} fails to accurately estimate body shape, leading to errors in clothed human reconstruction. The second data shows that ~\cite{SPIN,DecoMR,pymaf,pymaf-X} incorrectly recognizes the body as having bent legs when wearing loose coats, resulting in errors in clothed human reconstruction. In contrast, the results obtained by MuNet are the most reliable. Please zoom in to see the details.}
\label{ab2}
\end{figure}

The visual results in Figure~\ref{ab2} show that using existing methods to estimate 3D human meshes introduces pose and shape errors, leading to inaccuracies in clothed human reconstruction. The first data in the figure shows that existing advanced methods failed to accurately estimate the human shape, leading to errors in clothed human reconstruction. The second data in the figure demonstrates how existing methods incorrectly estimate body posture with loose clothing data, misinterpreting an upright posture as a bent-leg posture, leading to errors in clothed human reconstruction. In contrast, MuNet can accurately estimate body shape and pose under various types of clothing, resulting in more reasonable reconstruction outcomes.

\begin{table}
\caption{Quantitative results on guiding the reconstruction of 3D clothed human body models using different methods}
\resizebox{\linewidth}{!}{
\begin{tabular}{l|ccccc|ccccc|ccccc}
\hline
  & \multicolumn{5}{c|}{THuman 2.0} & \multicolumn{5}{c|}{CAPE} &\multicolumn{5}{c}{RenderPeople}\\
Prior Methods   & $\varepsilon_{cd}$ & $\varepsilon_{p2s}$ & $\varepsilon_{s2p}$ & $\varepsilon_{cos}$  & $\varepsilon_{l2}$& $\varepsilon_{cd}$  & $\varepsilon_{p2s}$& $\varepsilon_{s2p}$ & $\varepsilon_{cos}$  & $\varepsilon_{l2}$ & $\varepsilon_{cd}$  & $\varepsilon_{p2s}$ & $\varepsilon_{s2p}$& $ \varepsilon_{cos}$  & $ \varepsilon_{l2}$\\
\hline
SPIN~\cite{SPIN}      &4.141&3.508&4.774&0.3217	&0.6616	&3.820	&3.141 &4.499 &0.2372	&0.5002	&3.126&	2.666&3.585&0.2456&	0.5268\\

  GCMR~\cite{GCMR}      &3.642	&3.233	& 4.051&0.3021	&0.6291&3.660	&  2.967&4.354 &0.2285	&0.4854&2.874&2.465&3.282&0.2387	&0.5143\\
        DecoMR~\cite{DecoMR}      &	4.139&	3.396&4.881& 0.3138	&	0.6476&	3.828&3.082& 4.575 &0.2340	&0.4942&3.234&2.743&3.725& 0.2531&0.5403\\
    PyMAF~\cite{pymaf}       &4.113&3.483	&4.743&0.3197	&0.6584&	 3.483& 2.968&3.997&0.2238&0.4761	&3.141&2.674&3.608&0.2483&0.5317\\
      PyMAF-X~\cite{pymaf-X}      & 3.678	&	3.207&4.150	&0.2951&0.6180	&3.282&2.919 &3.644 &0.2274 &0.4845&	3.233&	2.727&3.739&0.2635&0.5591\\
      VirtualMarker~\cite{Virtual_Markers} &  3.136& 2.779&3.493& 0.2608&0.5567 &2.868&2.469 &3.267&0.1995 &0.4314&2.745&2.261&3.229&0.2246&0.4867\\
      MeshPose~\cite{meshpose}&3.314&2.876&3.752&0.2554& 0.5487&3.690&2.840&4.539&0.2183&0.4673&2.900&2.496& 3.304&0.2200&0.4829\\
     
    MuNet    &\textbf{2.141}	& \textbf{2.093}	& \textbf{2.189}		&\textbf{0.2026}& \textbf{0.4299} 	& \textbf{2.430}	& \textbf{2.393	}& \textbf{2.467}		&\textbf{ 0.1732}& \textbf{0.3247}& \textbf{1.345} & \textbf{1.236}&\textbf{1.455}&\textbf{0.1231}&\textbf{0.3049}\\
\hline
\end{tabular}
\label{tab:abl2}
}     
\end{table}

\subsubsection{Effectiveness of Loss Functions }
To verify that the accuracy of the recovered 3D mesh can be boosted by clothed 3D human reconstruction,
we train our method using different loss functions and test these models on the three benchmark datasets.
The quantitative results achieved by these models are shown in Table~\ref{tab:abl3}.
Without the ${\cal L}_{track}$ loss function, the performance of these metrics decreases compared to MuNet. Specifically, MPJPE, PA-MPJPE, and MVPE increased by 18.5mm, 11.8mm, and 16.8mm across the three datasets. Similarly, without the ${\cal L}_{cloth}$ loss function, MPJPE, PA-MPJPE, and MVPE increase by 13.2mm, 8.3mm, and 12.5mm, respectively.

\begin{table}
\caption{Comparisons of our method trained with different loss functions}
\resizebox{\linewidth}{!}{
\begin{tabular}{l|ccc|ccc|ccc}
\hline
  &\multicolumn{3}{c|}{THuman 2.0} & \multicolumn{3}{c|}{CAPE} &\multicolumn{3}{c}{Renderpeople}\\
 Loss Functions   & MPJPE & PA-MPJPE & MVPE &  MPJPE & PA-MPJPE & MVPE &MPJPE & PA-MPJPE & MVPE\\
\hline
     
    w/o ${\cal L}_{track}$  & 77.1&56.1 &78.6	&62.1&47.4&70.1&57.7&42.1&60.3\\
       
     w/o ${\cal L}_{cloth}$ &  70.1&49.1 &70.7   & 60.6	& 45.3&68.5 & 50.3& 40.6&56.8	\\
     w ${\cal L}_{track}$  and ${\cal L}_{cloth}$   & \textbf{45.7}  &  \textbf{36.1} & \textbf{49.9}	&\textbf{53.3} &\textbf{41.1} &\textbf{60.0} &\textbf{42.3} & \textbf{33.0}& \textbf{48.6}	\\
   
\hline
\end{tabular}
\label{tab:abl3}
}
\end{table}
\section{Discussion}
\textbf{Limitations.}
Although experimental results have validated the effectiveness of MuNet in leveraging the interplay between 3D human mesh recovery and 3D clothed human reconstruction, certain limitations remain. First, the current implementation of MuNet does not incorporate an iterative refinement between 3D human mesh recovery and 3D clothed human reconstruction during the inference phase, which could further enhance the performance of the two tasks. Additionally, like other 3D clothed human reconstruction methods, the current implementation of MuNet also lacks high-resolution facial details in its reconstructions.
This is primarily because the current implementation of MuNet employs the original SMPL model~\cite{SMPL} as the initial graph for deformation, whose resolution is inherently limited.

\textbf{Future Work.}
To address the limitations mentioned above, future work to improve MuNet can explore several promising directions. First, the inference process can be enhanced by introducing iterative refinement between 3D human mesh recovery and 3D clothed human reconstruction. Inspired by PaMIR~\cite{PaMIR}, such iterative interaction during inference may enable more consistent alignment between the underlying body shape and the reconstructed clothed surface, ultimately leading to improved reconstruction quality for both tasks. Second, the resolution of the initial graph used for deformation can be increased to better capture fine-grained details, especially in the facial region. This could be achieved by applying mesh interpolation techniques to upsample the SMPL topology. While higher-resolution graphs inevitably demand more GPU memory, one feasible solution is to follow the strategy of MeshCNN~\cite{meshcnn} by dividing the high-resolution graph into multiple subgraphs (e.g., eight parts), thus enabling efficient parallel processing and memory management during training and inference.

\section{Conclusion}
In this paper, we have proposed a unified framework that jointly addresses 3D human mesh recovery and 3D clothed human reconstruction from single images. By integrating both tasks into a single, closed-loop framework, we enable mutual guidance and joint optimization that were previously infeasible due to incompatible representations and isolated task designs.
Within this framework, we develop MuNet, a mutualistic network that employs a unified 2-manifold graph representation and an end-to-end graph convolutional network to progressively deform an initial graph into a 3D human mesh and a detailed 3D clothed human model. The mutualistic mechanism further allows reciprocal interaction between the tasks, improving performance in both mesh recovery and clothed reconstruction.
Extensive experiments on six benchmark datasets demonstrate that our MuNet achieves SOTA results on both tasks.
Therefore, we strongly recommend addressing these two tasks in a multi-tasking manner within a unified network. We hope that our work provides valuable insights for researchers working on both 3D human mesh recovery and 3D clothed human reconstruction.





\bibliographystyle{elsarticle-num} 
\bibliography{egbib}





\end{document}